\documentclass[10pt,twocolumn,letterpaper]{article}

\usepackage{wacv}
\usepackage{times}
\usepackage{epsfig}
\usepackage{graphicx}
\usepackage{amsmath}
\usepackage{amssymb}
\usepackage{caption}
\usepackage{balance}
\usepackage{subcaption}

\def\wacvPaperID{1296} 

\wacvfinalcopy 

\ifwacvfinal
\def\assignedStartPage{9876} 
\fi

\ifwacvfinal
\else
\fi

\begin{document}

\title{Attention-Guided Network for Iris Presentation Attack Detection}

\author{Cunjian Chen\\
Michigan State University\\
{\tt\small cunjian@msu.edu}
\and
Arun Ross\\
Michigan State University\\
{\tt\small rossarun@msu.edu}
}

\maketitle

\begin{abstract}
Convolutional Neural Networks (CNNs) are being increasingly used to address the problem of iris presentation attack detection.  In this work, we propose attention-guided iris presentation attack detection (AG-PAD) to augment CNNs with attention mechanisms.  Two types of attention modules are independently appended on top of the last convolutional layer of the backbone network. Specifically, the channel attention module is used to model the inter-channel relationship between features, while the position attention module is used to model inter-spatial relationship between features.  An element-wise sum is employed to fuse these two attention modules.  Further, a  novel hierarchical attention mechanism is introduced. Experiments involving both a JHU-APL proprietary dataset  and the benchmark  LivDet-Iris-2017 dataset suggest that the proposed method achieves promising results. To the best of our knowledge,  this is the first work that exploits the use of attention mechanisms in iris presentation attack detection. 
\end{abstract}

\section{Introduction}
Iris recognition systems are vulnerable to various types of presentation attacks (PAs), where an adversary presents a fabricated artifact or an altered biometric trait to the iris sensor in order to circumvent the system~\cite{Czajka:2018:PAD}. Commonly discussed attacks include cosmetic contacts, printed eyes, and artificial eye models~\cite{mtpad2018}. To address these challenges, convolutional neural networks (CNNs) are being increasingly used for  addressing the problem of  iris presentation attack detection (PAD)~\cite{HeLLLSH16, RaghavendraRB17, YadavWACV2018, DensePAD2019, PalaB17, KuehlkampPRBC19}, which is often formulated as a binary-classification task. The output of the network is a presentation attack (PA) score indicating whether the input image should be classified as ``live" or ``PA".\footnote{Sometimes the term ``bonafide" is used instead of ``live" and the term ``spoof" is used instead of ``PA"  in the biometrics literature.}  Existing work have shown remarkable performance on {\em known} or {\em seen}  presentation attacks where attacks encountered in the test set are observed in the train set~\cite{MenottiCPSPFR15}. However, degraded iris image quality can significantly affect  detection accuracy~\cite{YadavWACV2018}. Further, detecting {\em unknown} or {\em unseen}  presentation attacks remains a very challenging problem~\cite{LivDet2017}. 

Attention networks~\cite{xiaolongwang2017nonlocal, SENet2018} have emerged recently and they model the interdependencies between channel-wise features and/or spatial-wise features on CNN feature maps. The feature maps are produced by applying a series of convolution filters to a previous layer. The dimensionality of the feature maps is defined as channel$\times$height$\times$width. The channel corresponds to the number of convolution filters. The height$\times$width comprises of a spatial dimension. Channel-wise features derive from channel-wise convolution that operates along the direction of the channel of feature maps, whereas spatial-wise features derive from spatial-wise convolution that operates along the direction of the width and height of feature maps. These networks have been appropriated in the context of the face modality~\cite{Wang_2019_CVPR_Workshops,TwostreamAttention2020}, but have not been exploited by other modalities. Wang et al.~\cite{Wang_2019_CVPR_Workshops} proposed a multi-modal fusion approach by sequentially combining the spatial and channel attention modules to improve the generalization capability of face PAD systems. Chen et al.~\cite{TwostreamAttention2020} developed an attention-based fusion scheme that can effectively capture the feature complementarity from the outputs of two-stream face PAD networks. However, attention mechanisms have not been leveraged for use in {\em iris} presentation attack detection.  Further, effectively integrating such attention modules within the CNN architecture is yet to be systematically studied for presentation attack detection. 

\begin{figure*}[h!]
  \centering
    \includegraphics[width=0.8\textwidth]{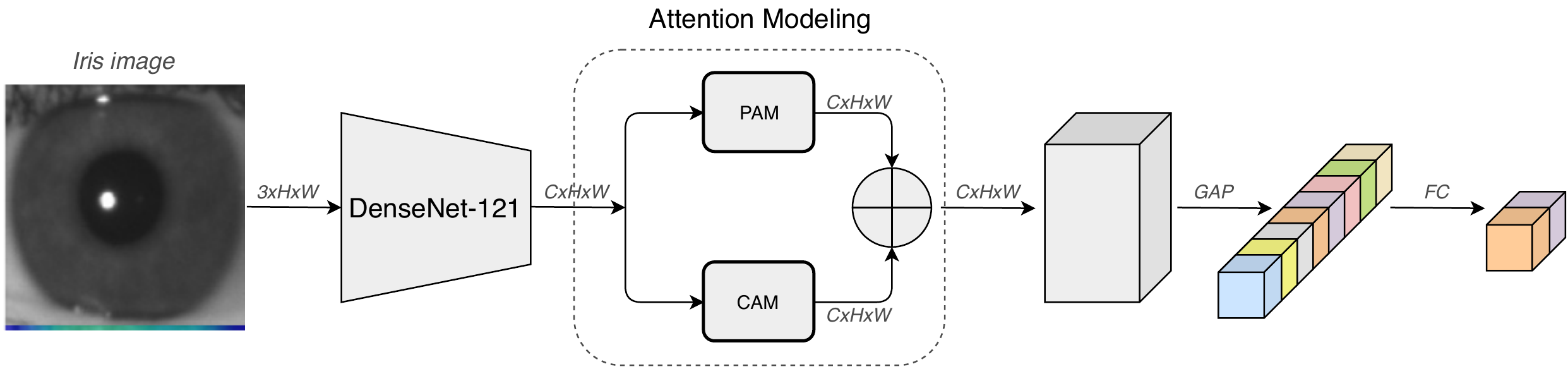}
    \caption{Flowchart depicting the proposed AG-PAD in this work. GAP and FC refer to the global average pooling and fully connected layers, respectively. The attention modeling process involves both position attention module (PAM) and channel attention module (CAM). $\oplus$ denotes element-wise sum. }
    \label{proposed_flowchart}
\end{figure*}

In this paper, we present an attention-guided iris presentation attack detection (AG-PAD) method that improves  the generalization capability of the existing PAD networks. This is due to the capability of attention mechanisms to model long-range pixel dependencies such that it can refine the feature maps to focus on regions of interests. Here, long-range dependencies are modeled via the receptive fields formed by a series of sequential convolutional operations. Given a set of convolutional feature maps obtained from a backbone network, a {\em channel-attention}  module and a {\em position-attention}  module are independently used to capture the inter-channel and inter-spatial feature dependencies, respectively. After that, the refined feature maps are input to convolutional blocks to extract more compact features. Finally, the outputs from these two attention modules are fused using an element-wise sum to capture complementary attention features from both channel and spatial dimensions. This is followed by global average pooling and softmax operations to compute the class probabilities  (``live" or ``PA").  The flowchart of the AG-PAD network is depicted in Figure~\ref{proposed_flowchart}.

The main contributions of this work are summarized here: 
\begin{itemize}
\item We propose an attention-guided PAD approach that can enhance the generalization capability of  iris presentation attack detection.
\item We extend the proposed PAD network by introducing hierarchical attention to attend to lower-layer feature representations.
\item We evaluate the proposed method on challenging datasets, which involves both unseen and unknown PAs.
\end{itemize}
Although attention modules have been exploited in the face PAD literature~\cite{Wang_2019_CVPR_Workshops}, our proposed method differs in several different ways: (a) we propose a parallel combination of these two different modules via an element-wise sum, instead of sequentially combining the attention modules; and (b) we construct a hierarchical attention to better attend to the low-layer feature representations. The proposed framework generalizes beyond iris modality and is applicable to face modality as well. 

The rest of the paper is organized as follows. Section~\ref{related_work} presents a literature survey on recent developments in iris presentation attack detection and attention mechanism. Section~\ref{proposed_method} describes the proposed attention-guided network used in this work, including the hierarchical attention network. Section~\ref{exp_results} discusses the result of the proposed method on both benchmark datasets and challenging proprietary datasets  and compare it with the state-of-the-art methods. Conclusions are reported in Section~\ref{conclusion}.

\section{Related Work}
\label{related_work}
In this section, we provide a brief discussion on existing iris PAD techniques that utilize convolutional neural network approaches and the applications of attention mechanisms in various computer vision tasks. \\
\indent \textbf{CNN-based Iris PAD:}  Development of Iris PAD approaches using CNN typically operate on either geometrically normalized~\cite{HeLLLSH16, RaghavendraRB17, DensePAD2019} or un-normalized iris images~\cite{PalaB17, mtpad2018, Yadav_2018_CVPR_Workshops, Hoffman_2018_CVPR_Workshops, KuehlkampPRBC19, YadavCR19}. This requires the use of an iris segmentation or an iris detection method as the preprocessing step~\cite{ChenR18}. He et al.~\cite{HeLLLSH16} proposed a multi-patch convolutional neural network (MCNN) approach that  densely samples iris patches from the normalized iris image. Each patch was independently fed into a convolutional neural network, where a final decision layer was used to fuse the outputs. On the other hand, Chen and Ross~\cite{mtpad2018} directly used the un-normalized iris image. Their proposed method simultaneously performs  iris localization and presentation attack detection. Kuehlkamp et al.~\cite{KuehlkampPRBC19} computed multiple binarized statistical image features (BSIF) representations of un-normalized iris images by exploiting different filter sizes and used these as inputs to the CNNs. An ensemble model was then used to  combine the outputs from the different BSIF representations. To handle the problem of unseen presentation attacks, Yadav et al.~\cite{YadavCR19} used relativistic average standard generative adversarial network (RaSGAN) to synthesize iris images in order to train a PAD system that can generalize well to ``unseen" attacks.  None of the aforementioned methods use attention modules to enhance the presentation attack detection performance.

\indent \textbf{ Attention Mechanism: } The use of an attention mechanism has been adopted in a variety of tasks such as image captioning~\cite{XuBKCCSZB15}, segmentation~\cite{DANET2019, HACCN2020}, classification and detection~\cite{xiaolongwang2017nonlocal, SENet2018, CBAM2018, XLWH19}. In addition, attention modules have also been used by generative adversarial networks (GANs) to allow for long-range pixel-dependency modeling for image generation task~\cite{SAGAN2019, Thermal2019}.  Generally, attention mechanisms can be coarsely divided into two types: the generation of channel attention module (CAM) and position attention module (PAM).\footnote{PAM is sometimes also referred to as spatial attention module.}  Hu et al.~\cite{SENet2018} proposed a novel architecture, termed ``Squeeze-and-Excitation" (SE) block, to explicitly model the interdependencies between channels. Their proposed SENet won the ILSVRC 2017 image classification competition and generalized well across challenging datasets. Though CAM and PAM can be independently integrated into existing network architectures, they can also be combined together to provide complementary attentions. Woo et al.~\cite{ParkWLK18} proposed a bottleneck attention module (BAM) to first compute the channel and spatial attentions along two separate network branches. Then, an element-wise sum was used to combine these two attention branches. The BAM was demonstrated to show consistent improvement in both classification and detection tasks. There are, of course, other ways to compute the channel and spatial attention modules. Fu et al.~\cite{DANET2019} utilized the self-attention mechanism to compute the PAM and CAM. The PAM and CAM was also combined via an element-wise sum. Their proposed dual attention network (DANet) achieved state-of-the-art performance in image segmentation.  Inspired by the works of~\cite{CBAM2018, DANET2019}, we argue that PAM and CAM are able to capture the long-range pixel dependencies along the spatial and channel dimensions, thereby refining the feature maps to focus on salient iris regions, such that it can extend the generalization capability of  iris PAD  solutions.  

\section{Proposed Algorithm}
Our proposed AG-PAD network aims to automatically learn discriminative features from the cropped iris regions that are relevant to presentation attack detection. However, there is a need for using attention mechanisms to further enhance the feature discrimination in both spatial and channel dimensions.  This will ensure that the network will focus more on salient regions during the backpropagation learning. The attention-based CNN model is designed by leveraging knowledge via transfer learning. Given a cropped iris image $I$ from a detection network~\cite{mtpad2018}, the feature maps are extracted by a backbone network $f$, which is formulated as:
\begin{equation}
A=f(I|\theta).
\end{equation}
Here,  $A\in\mathbb{R}^{C\times H \times W}$ denotes the feature maps of the last convolution layer, where $C$, $H$, $W$ are the number of channels, height, and width of the feature maps, respectively. $\theta$ is a set of parameters associated with the network. Without loss of generality, DenseNet121~\cite{Densenet2017} is chosen as our backbone network. 

\label{proposed_method}
\subsection{Position Attention Module}
Given output feature maps, $A\in\mathbb{R}^{C\times H \times W}$, obtained from a backbone network, it is first fed into two different convolution layers to produce feature maps $B\in\mathbb{R}^{C/r\times H \times W}$ and $C\in\mathbb{R}^{C/r\times H \times W}$, respectively. Here, $r$ is the reduction ratio. Then, these two feature maps are reshaped to $\mathbb{R}^{C/r\times N}$, where $N=H \times W$. After that, $C$ is transposed to $\mathbb{R}^{N\times C/r}$, and multiply with $B$ to obtain a feature map of size $\mathbb{R}^{N\times N}$. Finally, a softmax layer is applied to compute the position attention map $P\in\mathbb{R}^{N\times N}$ as follows:
\begin{equation}
P_{ij}=\frac{exp(C_i \cdot B_j)}{\sum_{i=1}^{N} exp(C_i \cdot B_j)},
\end{equation}
where, $C_i$ denotes the $i$-th row of $C$ and $B_j$ denotes the $j$-th column  of $B$. $P_{ij}$ is a probability value measuring the position dependency between $C_i$ and $B_j$, meaning that it can be considered as a weight to refine a pixel value in the spatial position of a feature map. 

On the other hand, the feature map $A$ is fed into another convolution layer to obtain a feature map $D\in\mathbb{R}^{C\times H \times W}$, which is later reshaped to $\mathbb{R}^{C\times N}$. After that, a matrix multiplication is performed between the reshaped $D$ and $P$ to obtain  $\mathbb{R}^{C\times N}$. This can be simply reshaped back to the original feature map dimension $\mathbb{R}^{C\times H \times W}$. The final refined output is obtained by:
\begin{equation}
M_{ij}=\alpha \sum_{k=1}^{N} (D_{ik} P_{kj} ) + A_{ij}
\end{equation}

\begin{figure}[h!]
  \centering
    \includegraphics[width=0.45\textwidth]{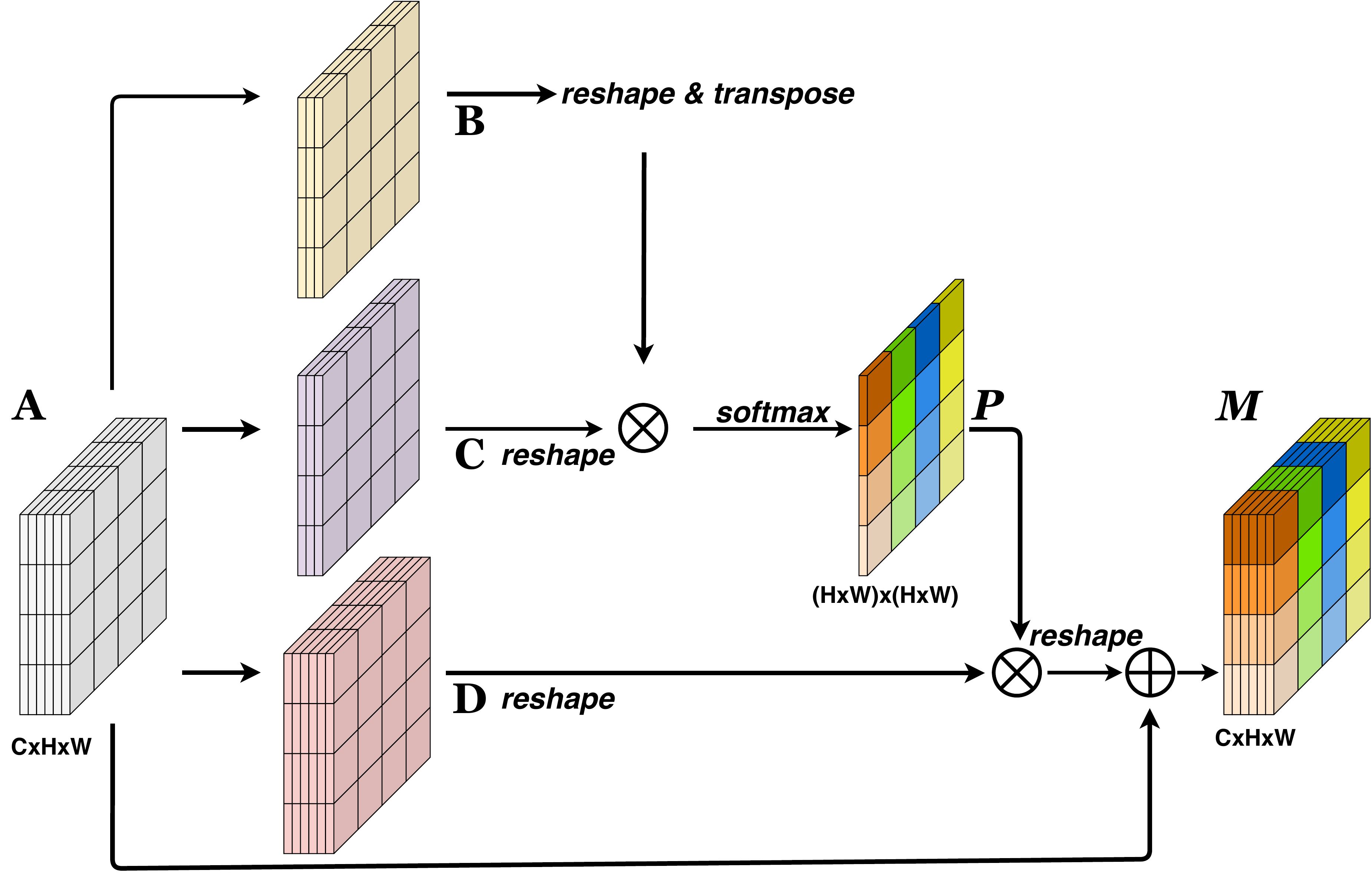}
    \caption{Flowchart depicting the computation of PAM in this work. $\oplus$ denotes element-wise sum and $\otimes$ denotes matrix multiplication.}
    \label{pam_flowchart}
\end{figure}

Visually, the procedures to compute the position attention map can be seen in Figure~\ref{pam_flowchart}.  

\subsection{Channel Attention Module}
Given an output feature map, $A\in\mathbb{R}^{C\times H \times W}$, obtained from a backbone network, it is first reshaped to $\mathbb{R}^{C\times N}$. Then, a matrix multiplication is performed between $A$ and the transpose of $A$, resulting in  $\mathbb{R}^{C\times C}$. The channel attention map $Q$ can be obtained by:

\begin{equation}
Q_{ij}=\frac{exp(A_i \cdot A_j)}{\sum_{i=1}^{C} exp(A_i \cdot A_j)},
\end{equation}
where, $A_i$ and $A_j$ are used to denote the $i$-th row and $j$-th column, respectively.  $Q_{ij}$ is a probability value measuring the channel dependency, meaning that it can be considered as a weight to refine a pixel value in the channel position of a feature map.  After that, the channel attention map $Q$ is multiplied with $A$, where the ensuing feature map is reshaped back to $\mathbb{R}^{C\times H \times W}$.  The final refined output form the channel attention map is computed by rescaling with $\beta$ and an element-wise summation with $A$:

\begin{equation}
M_{ij}= \beta \sum_{k=1}^{C} (Q_{ik} A_{kj})+ A_{ij}.
\end{equation}

\begin{figure}[h!]
  \centering
    \includegraphics[width=0.45\textwidth]{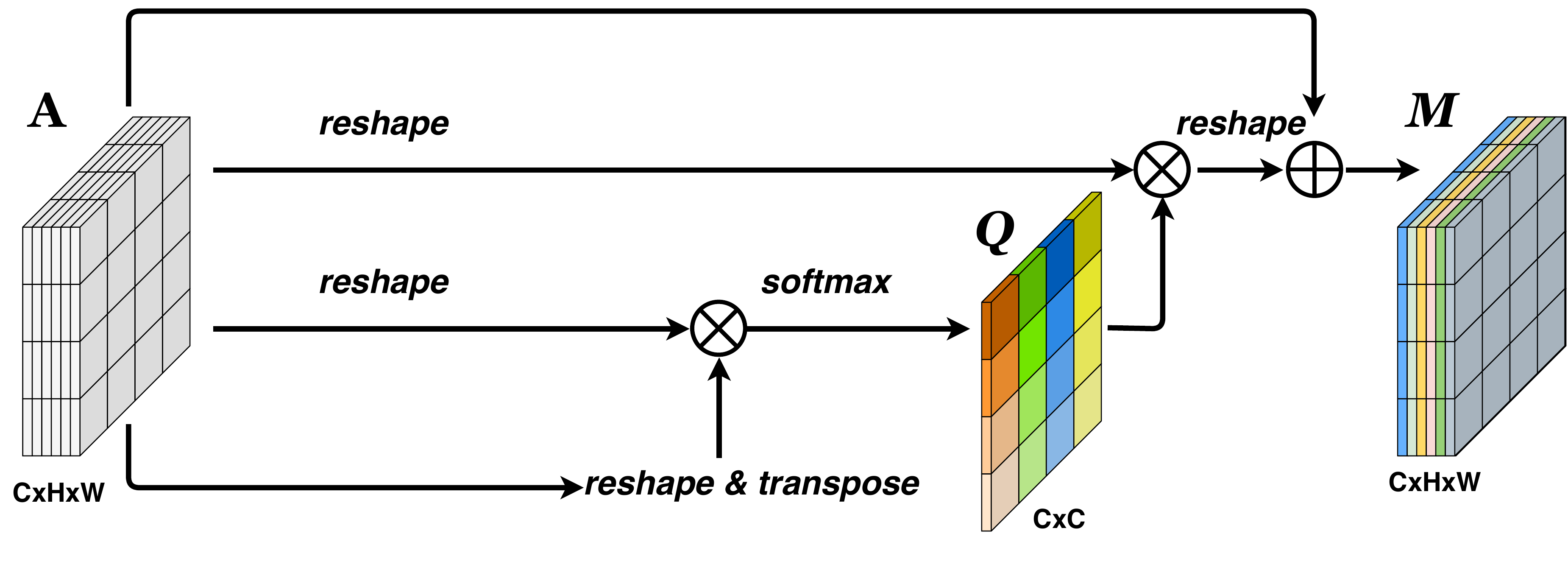}
    \caption{Flowchart depicting the computation of CAM in this work. $\oplus$ denotes element-wise sum and $\otimes$ denotes matrix multiplication.}
    \label{cam_flowchart}
\end{figure}

Visually, the procedures to compute the channel attention map can be seen in Figure~\ref{cam_flowchart}. 

\subsection{Attention-Guided PAD}
After utilizing the position attention module and the channel attention module to generate the refined outputs from the input feature maps $A$, the remaining task is to effectively fuse the attention information from the complementary modules. 

First, we consider to perform an element-wise sum to fuse the two attention modules. 
\begin{equation}
A'=M_p(A) \oplus M_q(A).
\end{equation}
Here, $M_p(A)$ and $M_q(A)$ are the refined feature maps after applying the PAM and CAM to the input feature maps $A$, respectively. 

Then, we also consider to sequentially combine these two attention modules. 
\begin{equation}
A'=M_q(A)
\end{equation}
\begin{equation}
A''=M_p(A').
\end{equation}
Here, $A'$ is the output after applying the channel attention module to the input feature maps $A$. $A''$ is the output after applying the position attention module to the input feature maps $A$.

In addition, a novel hierarchy-attention architecture  is proposed, which applies PAM and CAM to mid-level feature maps. Specifically, we have considered feature maps extracted from the \textit{conv3} block and \textit{conv4} block, which have dimensions of $28\times28\times512$ and $14\times14\times1024$, respectively. In addition, CAM is applied to the last convolutional block \textit{conv5} of size $7\times7\times1024$. Since the feature dimensions from individual convolutional blocks are different, max pooling is applied first to reduce both dimensions of $28\times28\times512$ and $14\times14\times1024$ to $7\times7\times 512$ and $7\times7\times1024$, prior to feature concatenation. We have demonstrated later that the hierarchy-attention architecture is more superior for iris presentation attack detection on low-quality samples due to its multi-scale representation.

\subsection{Implementation Details}\label{training_details}
Due to a limited number of training samples in existing iris presentation attack datasets, our PAD models are pre-trained on ImageNet~\cite{ImageNet2015}.  We fine-tune our models using the feature maps of the last convolutional layer. Unless specified, DenseNet121~\cite{Densenet2017} is used as the backbone network. The input spatial size is $224\times224$ pixels. Our models are trained for 50 epochs in total, using Adam optimizer with a learning rate of 0.0001. A mini-batch size of 32 is used during the training. We add two convolutional layers right after the attention layers, prior to the element-wise sum of the refined feature maps.  

To better generalize to unseen attacks,  extensive data augmentation is applied to populate the training dataset. A number of operations, including horizontal flipping, rotation, zooming, and translation are applied. It must be noted that sensor interoperability is implicitly addressed by gleaning iris samples from different datasets in the training phase. The iris detection module is implemented using the Darknet framework and the iris PAD is implemented using the  Keras framework. 

\section{Experimental Result}
\label{exp_results}
\subsection{Datasets and Metrics}
The proposed AG-PAD is evaluated on the datasets collected by Johns Hopkins University Applied Physics Laboratory (JHU-APL). The datasets are collected across two different sessions, which are termed as JHU-1 and JHU-2. To facilitate the comparison against state-of-the-art methods, the proposed method is also evaluated on the benchmark  LivDet-Iris-2017 datasets~\cite{LivDet2017}.  

\begin{figure}[h!]
  \centering
    \includegraphics[width=0.35\textwidth]{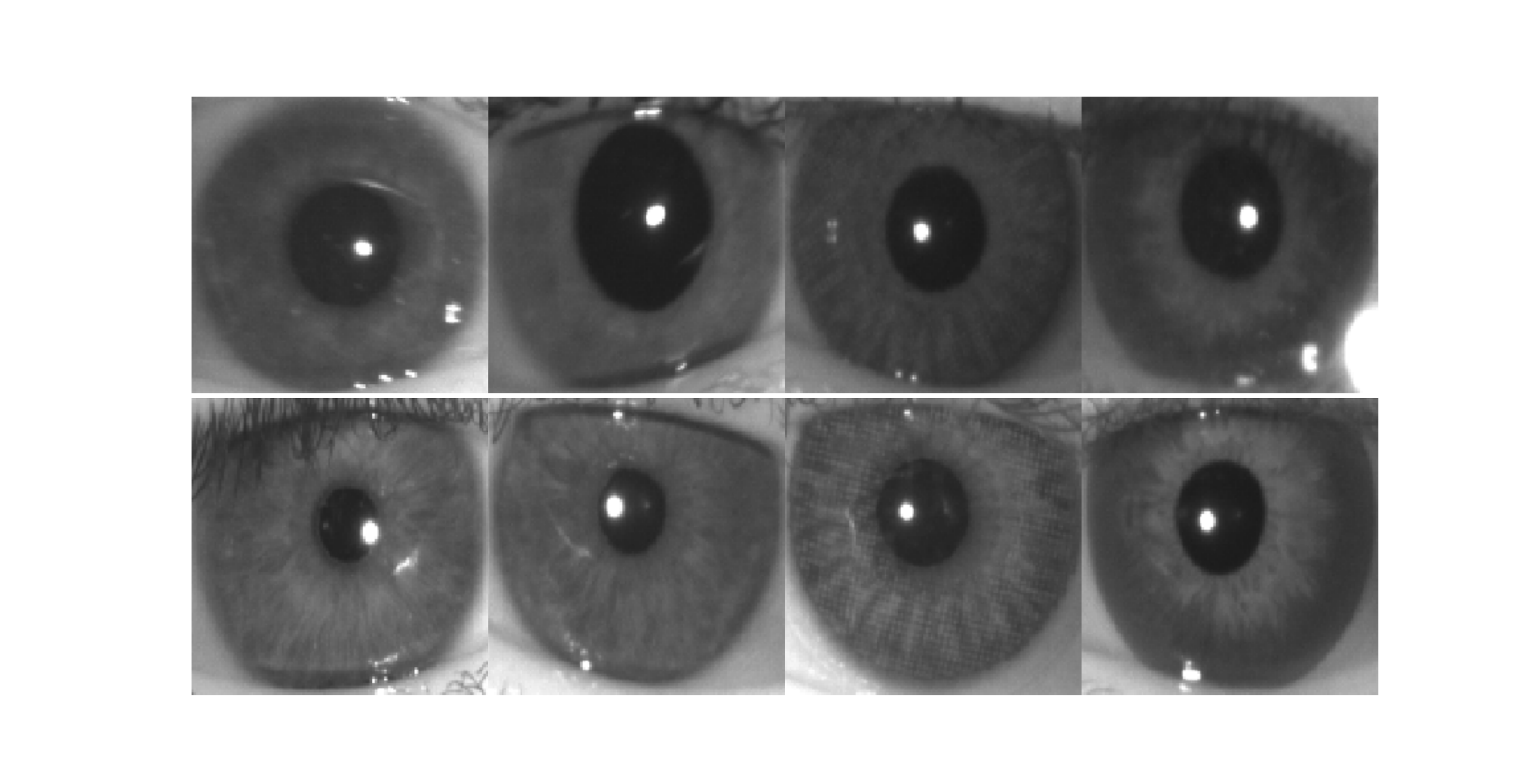}
    \caption{Examples of differences in image capture quality between JHU-1 and JHU-2 datasets. The images in the top and bottom rows are from JHU-1 and JHU-2 datasets, respectively. The left two columns denote the live samples, whereas the right two columns denote the PA samples.}
    \label{jhu_samples}
\end{figure}

\textbf{JHU-APL:} The training set consists of 7,191 live samples and 7,214 PA samples that are assembled  from a variety of datasets~\cite{LivDet2017}. JHU-1 consists of 1,378 live samples and 160 PA samples. Types of PAs in JHU-1 include colored contact lenses and Van Dyke/Doll fake eyes.  JHU-2 consists of 1,368 live samples and 227 PA samples.  Types of PAs in JHU-2 include colored contact lenses and Van Dyke/Doll fake eyes. The colored contact lens in JHU-2 can be further divided into Air Optix colored contact lenses, Acuvue Define colored contact lenses, and Intrigue partial coverage lenses (some examples are shown in Figure~\ref{evaluation_by_visual}). The image quality between JHU-1 and JHU-2 were observed to be different. Samples collected for JHU-2 have improved capture quality by minimizing variations such as blur and reflection (see Figure~\ref{jhu_samples}). The experiments were conducted in a cross-dataset setting, where training and test subsets are from different datasets. 

\textbf{LivDet-Iris-2017:} For LivDet-Iris-2017-Warsaw, the training subset consists of 2,669 print samples and 1,844 live samples. The test subset contains both known spoofs and unknown spoofs. The known-spoofs subset includes 2,016 print samples and 974 live samples, while the unknown-spoofs subset  includes 2,160 print samples and 2,350 live samples. For LivDet-Iris-2017-ND,  the training subset consists of 600 textured contact lenses and 600 live samples. The testing subset is split into known spoofs and unknown spoofs. The known-spoofs subset includes 900 textured contact lenses and 900 live samples. The unknown-spoofs subset  includes 900 contact PAs, where the types of contact lenses are not represented in the training set, and 900 live samples. 

\textbf{Evaluation Metrics:} To report results on JHU-APL datasets, we use True Detection Rate (TDR: rate at which PAs are correctly classified) and False Detection Rate (FDR: rate at which live images are misclassified as PAs),\footnote{FDR defines how many `live'' images are misclassified as ``PA''. TDR at 0.2\% FDR was used to demonstrate the performance of this algorithm in practice.} along with the Receiver Operating Characteristic (ROC) curve. To report results on LivDet-Iris-2017, we use  the same evaluation metrics as outlined in the LivDet-Iris-2017 competition: (a) BPCER is the rate of misclassified live images (``live'' classified as ``PA''); and (b) APCER is the rate of misclassified PA images (``PA'' classified as ``live''). Note that FDR is the same as BPCER and TDR equals (1-APCER). Evaluations on JHU-APL and LivDet-Iris-2017 datasets follow the same training protocol described in Section~\ref{training_details}. The difference lies in what datasets are used for training and testing. 

\subsection{Evaluation on JHU-APL Datasets}
\begin{figure}[h!]
    \centering
    \begin{subfigure}[b]{0.5\textwidth}
        \includegraphics[width=\textwidth]{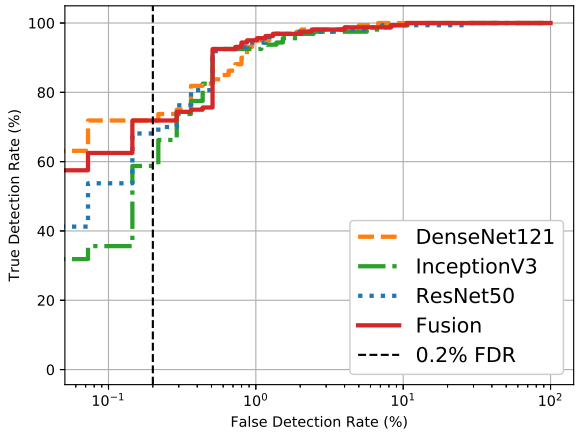}
        \caption{JHU-1}
        \label{ROC_Backbone_Comparison_GCT1}
    \end{subfigure}
    ~ 
    \begin{subfigure}[b]{0.5\textwidth}
        \includegraphics[width=\textwidth]{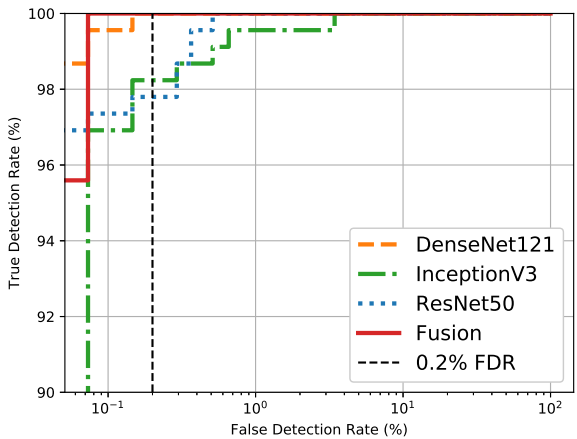}
        \caption{JHU-2}
        \label{ROC_Backbone_Comparison_GCT2}
    \end{subfigure}      
    \caption{Evaluation of the proposed AG-PAD method with different backbone networks on JHU-APL datasets.}
    \label{ROC_Backbone_Comparison}
\end{figure}

The purpose of JHU-APL datasets is to evaluate the generalizability of the iris PAD solutions in practical applications.  In addition to the DenseNet121 backbone network, we  also investigated the use of other backbones, viz.,  InceptionV3 and ResNet50.  The reason for testing various different backbone networks is to showcase the effectiveness of attention modules regardless of the choice of the  backbone network.  Moreover, an ensemble of multiple backbones may further boost PAD performance.  As can been seen from Figure~\ref{ROC_Backbone_Comparison}, the obtained TDR accuracy at 0.2\% FDR on JHU-1 dataset is not high. This is due to the degraded iris image quality originating from blur, reflections and glasses, to name a few.  After the image quality was improved, the proposed method achieved significantly better performance on the JHU-2 dataset (see Figure~\ref{ROC_Backbone_Comparison}).  Among all the three evaluated backbone networks, DenseNet121 obtains the best performance (see Table~\ref{ROC_Backbone_Comparison_table}). Though the average-score fusion of all three backbone networks does not improve performance on the JHU-1 dataset, it shows  improved performance on the JHU-2 dataset under a lower FDR (e.g., 0.1\% FDR). 

\begin{table}[h]
\caption{Evaluation of the proposed AG-PAD method with different backbone networks on JHU-APL datasets.  TDR at 0.2\% FDR was used to report the PAD accuracy.}
\begin{center}
  \begin{tabular}{ | c | c | c |  c | c}
    \hline
     & \textbf{JHU-1} & \textbf{JHU-2}  \\ \hline
    DenseNet121 & 71.87 &  100.0 \\ \hline
    InceptionV3  &  58.75 & 98.23 \\ \hline
    ResNet50 &  68.12 & 97.79 \\ \hline
    Fusion &  71.87&  100.0\\ 
    \hline
  \end{tabular}
\end{center}
\label{ROC_Backbone_Comparison_table}
\end{table}

\begin{figure}[h!]
  \centering
    \includegraphics[width=0.5\textwidth]{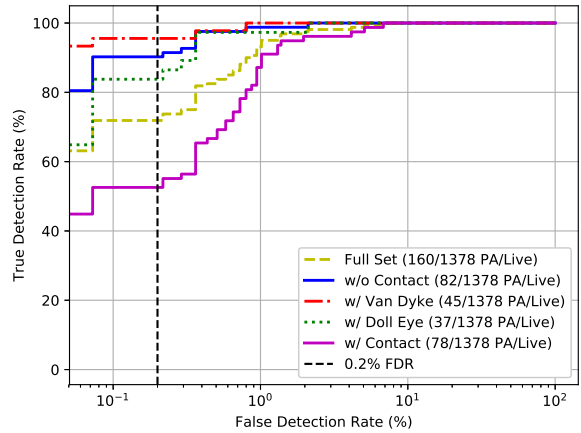}
    \caption{Evaluation of the proposed AG-PAD method on JHU-1 dataset with respect to different presentation attacks. The used backbone network is DenseNet121.}
    \label{evaluation_by_pa}
\end{figure}

The iris PAD performance was observed to vary with the nature of the presentation attack. While a presentation attack with patterned contact lenses results in only a subtle texture change to the iris region, those based on artificial eye models can extend beyond the iris  region and change the iris appearance significantly. Hence, the latter is much easier to detect, as evidenced by a higher true detection rate at 0.2\% FDR (see Figure~\ref{evaluation_by_pa}). This highlights the necessity to develop more effective PAD solutions for the cosmetic contact PA. 

\subsection{Architecture Design}
\begin{figure}[h!]
    \centering
    \begin{subfigure}[b]{0.5\textwidth}
        \includegraphics[width=\textwidth]{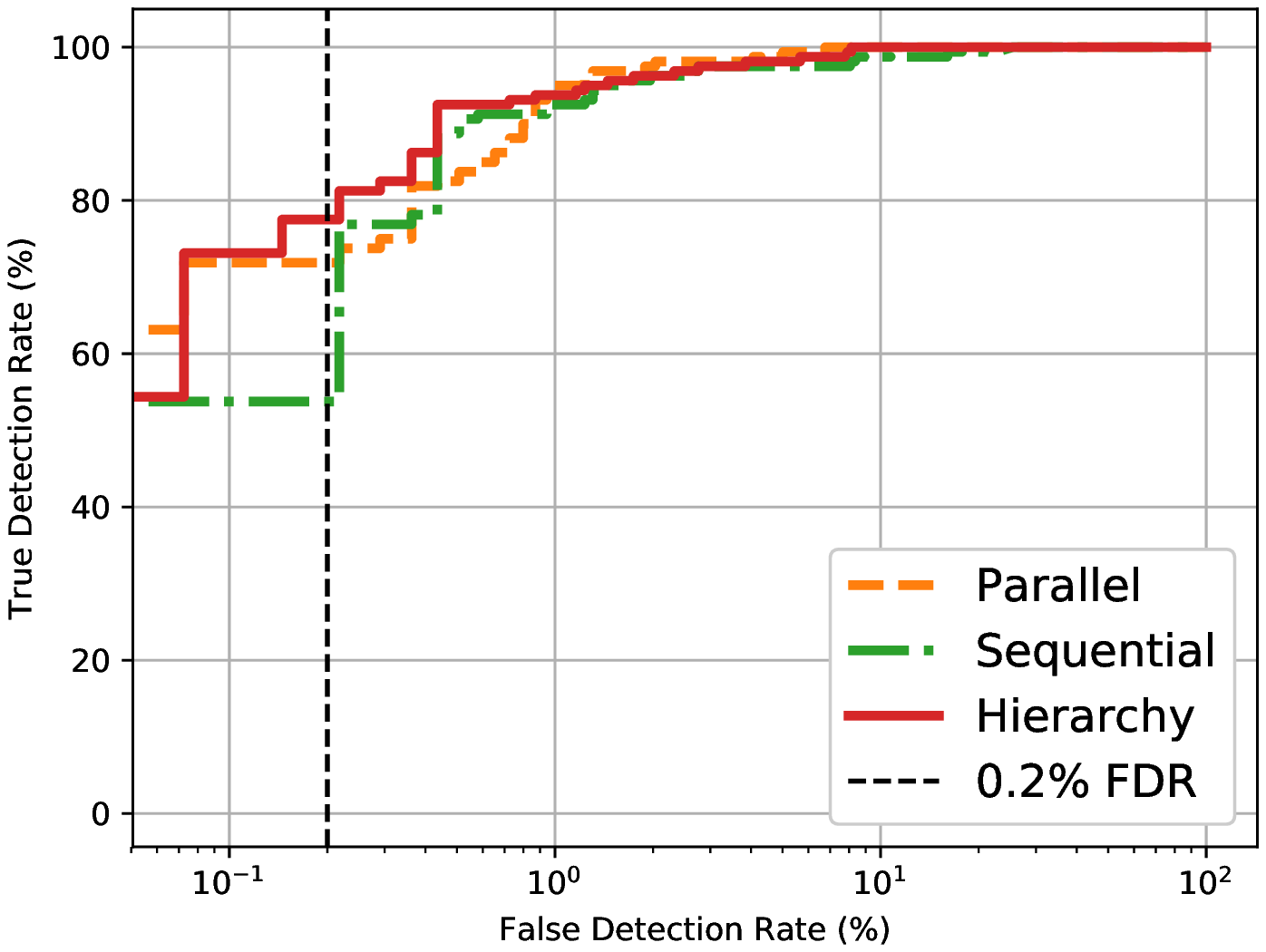}
        \caption{JHU-1}
        \label{arch_gct1}
    \end{subfigure}
    ~ 
    \begin{subfigure}[b]{0.5\textwidth}
        \includegraphics[width=\textwidth]{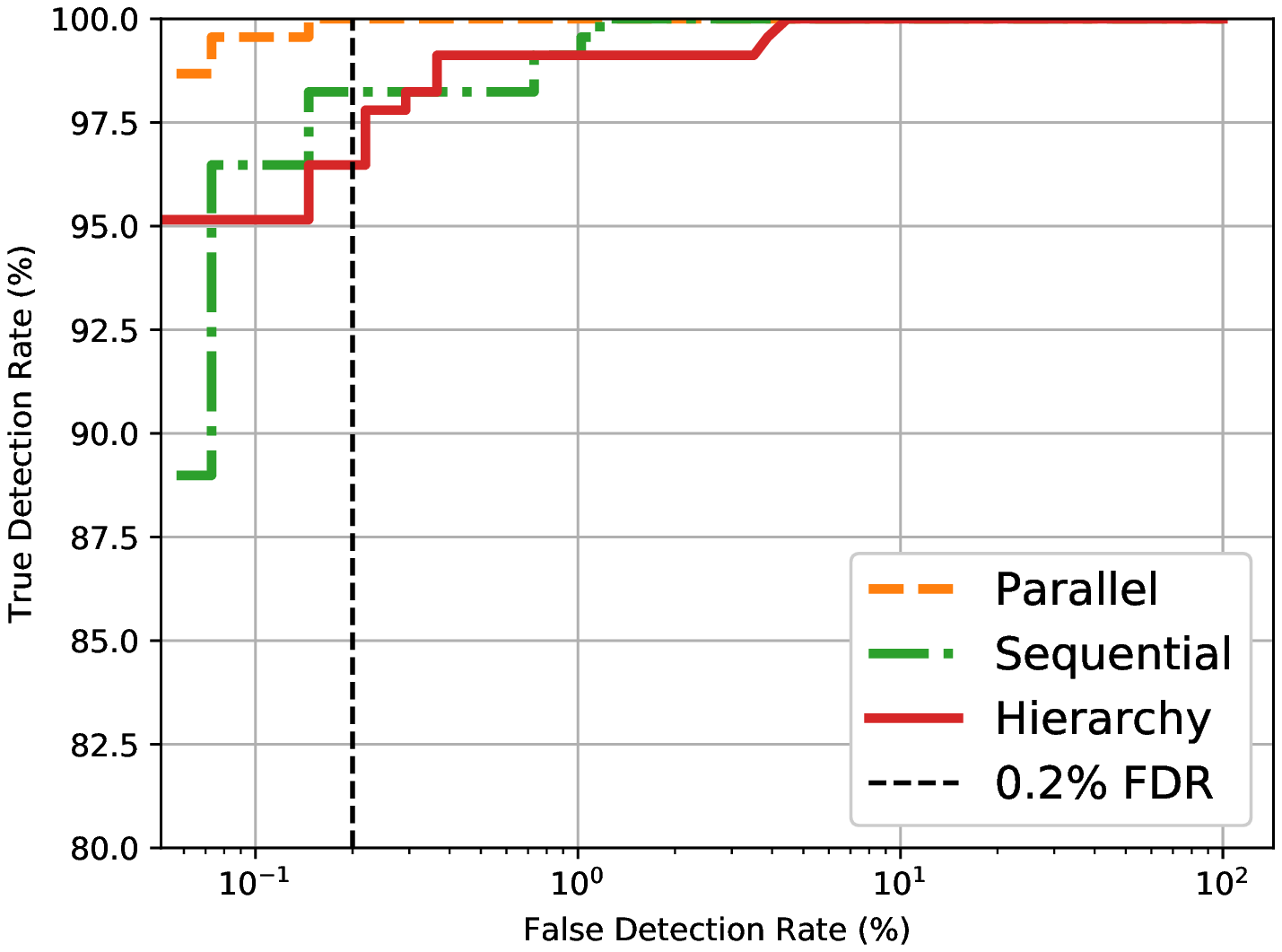}
        \caption{JHU-2}
        \label{arch_gct2}
    \end{subfigure}      
    \caption{Evaluation of the proposed methods with different architectures on JHU-APL datasets.}
    \label{arch_evauation}
\end{figure}

Previously, we mentioned three different ways to combine the PAM and CAM.  In this section, we show the evaluation results for all the three architectures on the JHU-APL datasets as well. As can be seen from Figure~\ref{arch_evauation}, the parallel combination (AG-PAD) obtains better performance than the sequential combination of CAM and PAM on both JHU-1 and JHU-2 datasets. The parallel combination obtains 71.87\% and 100.0\% TDRs at 0.2\% FDR on JHU-1 and JHU-2 datasets, respectively. The sequential combination obtains 53.75\% and 98.23\% TDRs at 0.2\% FDR  on JHU-1 and JHU-2 datasets, respectively. The hierarchy architecture, on the other hand, shows more promising results on the challenging JHU-1 dataset that has seen more image quality degradations. 

\subsection{Ablation Study}
To demonstrate the effectiveness of different attention modules, an ablation study was conducted using both the JHU-1 and JHU-2 datasets. In particular, four different variants are considered: \textit{w/o Attention, w/ PAM, w/ CAM,} and \textit{w/ PAM-CAM.} Baseline \textit{w/o Attention} refers to the results obtained by retraining the PAD network without any attention module. \textit{w/ PAM} and \textit{w/ CAM} refer to the results obtained by retraining the PAD network with appended PAM and CAM, respectively. Finally, \textit{w/ PAM and CAM} refers to the results obtained by augmenting the PAD network with both PAM and CAM. 

\begin{figure}[h!]
    \centering
    \begin{subfigure}[b]{0.5\textwidth}
        \includegraphics[width=\textwidth]{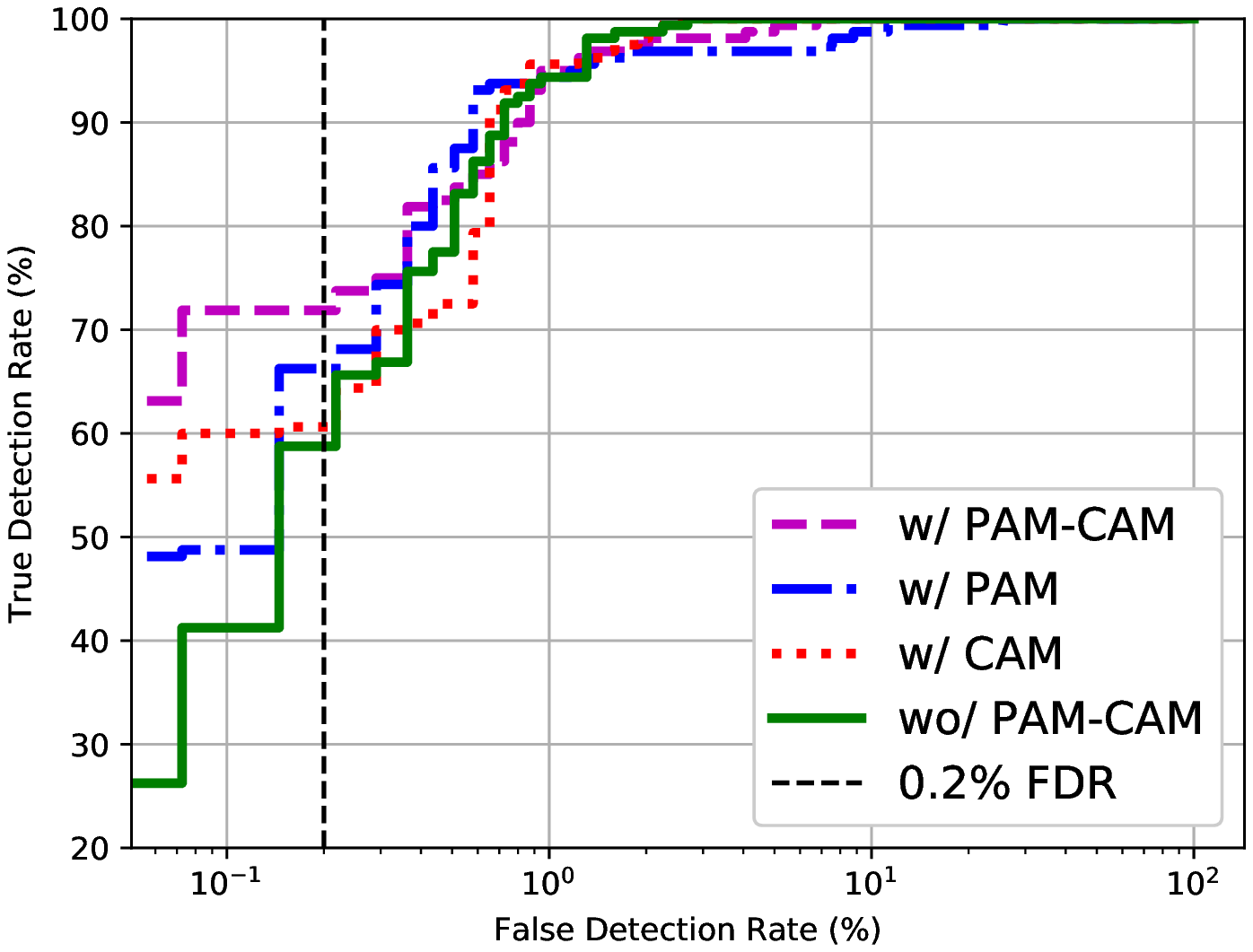}
        \caption{JHU-1}
        \label{ablation_gct1}
    \end{subfigure}
    ~ 
    \begin{subfigure}[b]{0.5\textwidth}
        \includegraphics[width=\textwidth]{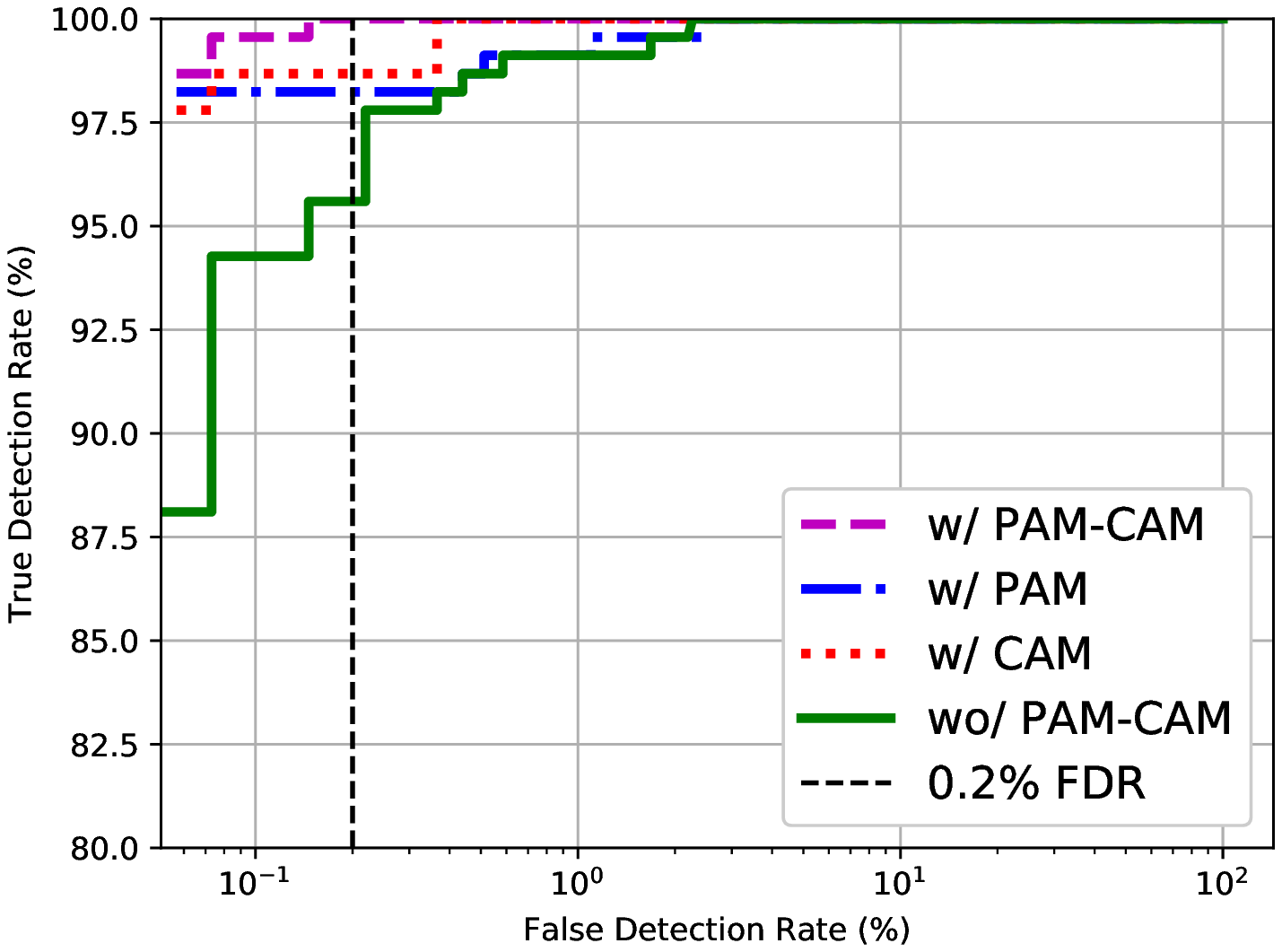}
        \caption{JHU-2}
        \label{ablation_gct2}
    \end{subfigure}      
    \caption{Ablation study of iris PAD performance with different attention mechanisms on the JHU-APL datasets. }
    \label{ablation_study}
\end{figure}

As can be seen in Figure~\ref{ablation_study}, the use of attention modules significantly improves the accuracy over the baseline. Without the attention modules, the baseline only gives 58.75\% TDR at 0.2\% FDR on JHU-1. After integrating both attention modules, the results improved to  71.87\% TDR (see Table~\ref{ablation_gct}).  A similar trend was observed in JHU-2 dataset. Further, comparing to other state-of-the-art attention modules~\cite{ParkWLK18,CBAM2018,XLWH19}, the combination of PAM and CAM achieves the best performance.  This justifies the significance of using the PAM and CAM attention modules in iris presentation attack detection. 

\begin{table}[h]
\caption{The TDR at 0.2\% FDR when using the PAM and CAM attention modules with DenseNet121 as the backbone network on the JHU-1 and JHU-2 datasets. The ablation study of the used attention module and its comparison against other attention modules can be seen here. }
\begin{center}
  \begin{tabular}{ | c | c | c |  c | }
    \hline
     & \textbf{JHU-1} & \textbf{JHU-2}  \\ \hline
    w/o Attention & 58.75  & 95.59\\ \hline
    w/ PAM   & 66.25  & 98.23\\ \hline
    w/ CAM & 60.62   & 98.67\\ \hline
    w/ PAM and CAM  & \textbf{71.87} & \textbf{100.0} \\ \hline
    BAM~\cite{ParkWLK18} & 68.75 & 97.35\\ \hline    
    CBAM~\cite{CBAM2018} & 70.0 & 97.79 \\  \hline  
    GC~\cite{XLWH19} & 66.87 & 92.95  \\ 
    \hline
  \end{tabular}
\end{center}
\label{ablation_gct}
\end{table}


\subsection{Visualization}
\begin{figure}[h!]
  \centering
    \includegraphics[width=0.45\textwidth]{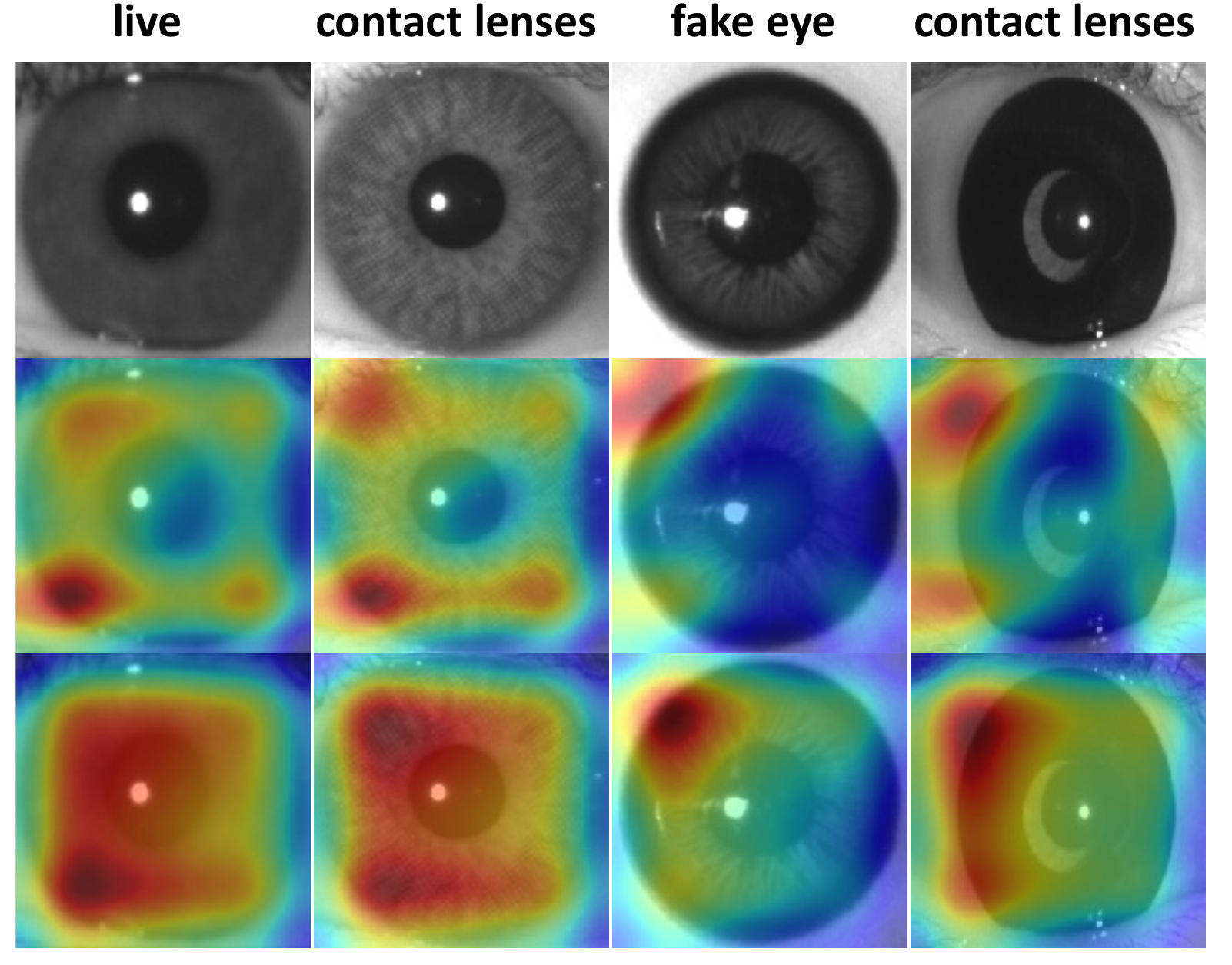}
    \caption{Visualization using Grad-CAM before and after the integration of attention maps for live, contact lenses and fake eye iris images, respectively. The second row is the result before the use of attention module and the third is the result after the use of attention module.}
    \label{attention_visualization}
\end{figure}

To further identify the important regions of an iris image that is used to render a PA decision, gradient-weighted class activation mapping (Grad-CAM)~\cite{Selvaraju_grad_cam} is utilized to generate the visualizations before and after the application of the attention modules in Figure~\ref{attention_visualization}. Here, Grad-CAM is used to calculate the gradient of the presentation attack detection score with respect to the feature maps to measure pixel importance. It is evident that the use of attention modules has enabled the network to shift the focus on to the circular iris. By observing the activation maps generated for both live and cosmetic contact samples (see Figure~\ref{attention_visualization}), the application of attention modules has forced the network to attend to the circular iris region in order to make the final decision. This is consistent with our intuition that iris texture positioned beyond the pupil region plays a much more significant role for presentation attack detection. 

\begin{figure}[h!]
  \centering
    \includegraphics[width=0.5\textwidth]{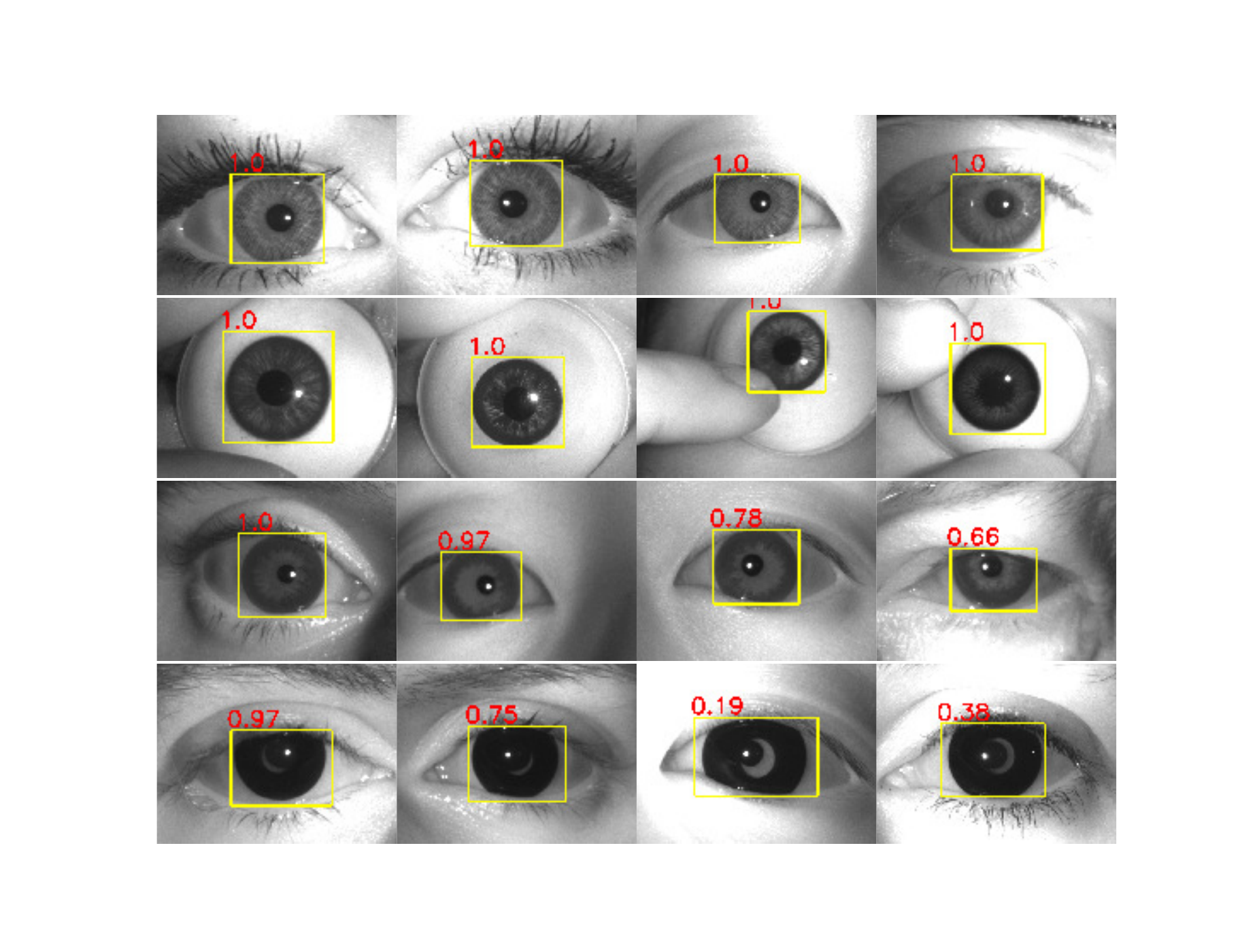}
    \caption{Evaluation of the proposed AG-PAD method on JHU-2 dataset with known and unknown presentation attacks. The first and second rows show the known PA results for Air Optix colored contact lenses and Van Dyke fake eye, respectively. The third row shows the unknown PA results for Acuvue Define colored contact lenses. The last row shows the unknown PA results for Intrigue partial coverage lenses with both successful and failure cases. }
    \label{evaluation_by_visual}
\end{figure}

In addition, we also visualize the PAD results for both known PAs and unknown PAs in Figure~\ref{evaluation_by_visual}. The bounding boxes are obtained from a pre-trained iris detection network.  Apparently, it is much more challenging to perform PAD on unknown PAs, such as Acuvue Define colored contact lenses and Intrigue partial coverage lenses (see Figure~\ref{evaluation_by_visual}). The unknown PA scores are observed to be much lower than the known PA scores. 

\subsection{Comparison with State-of-the-art Methods}
We also compare the proposed method against state-of-the-art methods evaluated on the LivDet-Iris-2017~\cite{LivDet2017}. Results of three algorithms, that participated in the competition, under both known presentation attacks and unknown presentation attacks were used. According to the evaluation protocol, a threshold of 0.5 was used to calculate the APCER and BPCER. 

As indicated in Table~\ref{livdet_results}, error rates are much higher for unknown presentation attacks on both Warsaw and Notre Dame datasets. Nevertheless, the proposed method achieves remarkable performance for both known and unknown presentation attacks. The proposed AG-PAD method achieves 0.09\% APCER and 0\% BPCER for unknown PAs in the Warsaw dataset. This is expected since the used dataset is limited to the print attacks only. 

\begin{table}[h!]
\scriptsize
\caption{Evaluation of the proposed method on the LivDet-Iris-2017 dataset. Both known/unknown results were reported whenever available.}
\begin{center}
  \begin{tabular}{ | c | c | c |  c | c | c |}
    \hline
    & \multicolumn{2}{c|}{\textbf{Warsaw}} & \multicolumn{2}{c|}{\textbf{Notre Dame}}   \\ \hline
     Algorithm & \textbf{APCER (\%)} & \textbf{BPCER (\%)} & \textbf{APCER (\%)} & \textbf{BPCER (\%)} \\ \hline
    CASIA~\cite{LivDet2017} & 0.15/6.43 & 5.74/9.78  & 1.56/21.11 & 7.56 \\ \hline
    Anon1~\cite{LivDet2017} & 0.4/11.44 & 2.77/6.64 & \textbf{0}/15.56 & 0.28\\ \hline
    UNINA~\cite{LivDet2017} & 0.1/\textbf{0} & 0.62/20.64 & 0.89/50 & 0.33\\ \hline
   Proposed  & \textbf{0.09}/1.34 & \textbf{0}/\textbf{0}  & 0.11/\textbf{8.33}& \textbf{0.22}\\ \hline
  \end{tabular}
\end{center}
\label{livdet_results}
\end{table}

\section{Conclusions}
\label{conclusion}
This paper proposes a novel attention-guided CNN framework for iris presentation attack detection. The proposed AG-PAD method utilizes attention-guided feature maps extracted by the channel attention module and the position attention module to regularize the network to focus on salient iris regions, thereby improving the generalization capability of iris PAD solutions. Experiments on several datasets indicate that the proposed method is effective in detecting both known and unknown presentation attacks. Future work would involve experiments on explainable attention mechanisms in iris PAD. 

\section{Acknowledgment}
This research is based upon work supported in part by the Office of the Director of National Intelligence (ODNI), Intelligence
Advanced Research Projects Activity (IARPA), via IARPA RD
Contract No. 2017 - 17020200004. The views and conclusions
contained herein are those of the authors and should not be interpreted as necessarily representing the official policies, either expressed or implied, of ODNI, IARPA, or the U.S. Government.

\balance
{\small
\bibliographystyle{ieee_fullname}
\bibliography{egbib}
}

\end{document}